\RequirePackage[T1]{fontenc}
\documentclass[letterpaper,journal]{IEEEtran}
\usepackage{amsmath,amssymb,bm}
\usepackage{graphicx}

\usepackage{booktabs,tabularx,multirow,array,ragged2e,makecell,threeparttable}
\usepackage[caption=false,font=normalsize,labelfont=sf,textfont=sf]{subfig}
\usepackage{textcomp,stfloats,url,cite,microtype,xcolor,enumitem}
\begin{document}
\title{Towards Privacy-Preserving Thermal Human Perception from Dataset to Deployment}
\author{Kailai~Sun, Yue~Cui, Xinwei~Wang, Zijie~Gao, Baoshen~Guo, Zhiqing~Hong, Irfan~Qaisar, Xu~Yang, Yueting~Li, Dian~Fan, and Qianchuan~Zhao%
\thanks{Kailai Sun, Yue Cui and Xinwei Wang contributed equally to this work.}%
\thanks{Kailai Sun and Baoshen Guo are with the Urban Mobility Lab, Massachusetts Institute of Technology, Cambridge, MA 02139, United States and the Singapore-MIT Alliance for Research and Technology Centre (SMART), Massachusetts Institute of Technology, Singapore 138602 (e-mail: skl24@mit.edu).}%
\thanks{Yue Cui is with Chongqing General Hospital, Chongqing University, Chongqing 401147, China (e-mail: cuiyue\_medicine@cqu.edu.cn).}%
\thanks{Xinwei Wang, Yueting Li, Dian Fan are with Hong Kong Xsmart Century Technology Co., Ltd., Hong Kong, China.}%
\thanks{Zijie Gao is with NUS-ISS, National University of Singapore, 119615, Singapore.}%
\thanks{Zhiqing~Hong is with Information Hub, The Hong Kong University of Science and Technology (Guangzhou), Guangzhou, China.}
\thanks{Irfan~Qaisar, Xu~Yang and Qianchuan~Zhao are with the Center for Intelligent and Networked Systems, Department of Automation, BNRist, Tsinghua University, Beijing 100084, China.}}
\markboth{IEEE Transactions}{Sun \MakeLowercase{\textit{et al.}}: Indoor RGB-T}
\maketitle
\begin{abstract}
Indoor human perception is the foundation for smart home and healthcare. However, RGB cameras raise privacy concerns in private indoor environments, while low-resolution thermal cameras provide limited spatial and semantic information. Large-scale indoor thermal datasets and real-world deployments remain limited. Here, we introduce Indoor RGB-T, a large-scale paired RGB--thermal dataset for indoor human perception. The dataset covers 20 indoor rooms and contains 90,576 $80\times62$ thermal frames, 87,016 temporally matched RGB--thermal pairs, and 115,557 annotated human bounding boxes. We establish benchmarks for robust thermal occupant-state detection with 0.75 mAP and occupancy estimation with an accuracy of 0.847 in unseen rooms. We further evaluate RGB-to-thermal and thermal-to-RGB translation, achieving a temperature MAE of $0.654\,^{\circ}\mathrm{C}$ and an $R^2$ of 0.923. Finally, we deploy the thermal perception model on a commercial sleep-monitoring system in the real world. Field testing with 100 participants and deployment of approximately 1,000 devices further demonstrate its practical feasibility. We identify practical challenges that motivate future research directions, including environmental thermal interference, sensing range and multi-person occlusion. The code, trained models and partial dataset are available at https://github.com/kailaisun/RGBT and https://huggingface.co/skl24/RGBT.
\end{abstract}

\section{Introduction}
\label{section1}

Indoor human perception \cite{schweiker2020review} is an important component of smart buildings, ambient intelligence, healthcare monitoring, and human-centered AI and automation. Traditionally, RGB cameras have been the mainstream because they provide high spatial resolution and rich semantic information, advancing the deployment of state-of-the-art (SOTA) deep learning models for indoor person detection \cite{sun2022fusion}, pose estimation \cite{8765346}, Re-ID \cite{11673055}, and activity recognition \cite{javadi2025graph}. Despite these technological advances, indoor human perception remains strongly constrained by privacy and ethical concerns. Deploying RGB cameras in private indoor environments, such as bedrooms, patient rooms, and residential spaces, raises severe privacy concerns \cite{liu2025mosaic,choi2021application}.  As a result, research attention has increasingly shifted toward privacy-preserving sensing modalities \cite{sun2020review}. While alternative non-intrusive sensing technologies \cite{caballero2024analysis} such as Wi-Fi \cite{11006977}, LiDAR \cite{11595253}, mmWave radar \cite{wu2023non}, Bluetooth, and passive infrared sensors provide better privacy protection, they generally suffer from coarse spatial resolution and less semantic information. These challenges limit the ability to capture and measure fine-grained indoor human states \cite{11154968,sun2020review}.

Unlike conventional RGB imaging, thermal infrared imaging can capture human-related information \cite{11154968} while significantly reducing the visibility of facial and personal features \cite{alsafery2023sensing}. Low-resolution thermal cameras have therefore emerged as promising alternatives for indoor human detection \cite{he2023smart}, localization \cite{geng2022lightweight}, pose and activity recognition \cite{lupion20243d}, environmental monitoring\cite{7447775,jiaoyu2025real}, and person identification \cite{11154968}. For example, previous studies \cite{11154968} have proposed an infrared-based person Re-ID method with an improved KCF tracking algorithm using thermal array sensors.

Despite increasing interest in indoor thermal sensing, several challenges remain for indoor human perception. First, existing RGB--thermal datasets and perception benchmarks are mainly designed for outdoor pedestrian detection, surveillance, and autonomous driving, while large-scale paired data collected in indoor environments remain limited. Indoor scenes contain different geometric layouts, human states, occlusion patterns, and thermal interference. Unlike thermal-based human perception in outdoor environments, human body temperature can be close to that of surrounding objects in indoor environments, such as beds, bedding, electronic devices, and other heat-emitting sources, resulting in severe thermal crossover. This low thermal contrast, together with background clutter and the inherently low spatial resolution of compact thermal cameras, makes reliable indoor human perception difficult.

Second, the capability of modern computer vision models to extract human location, state, and number directly from native low-resolution thermal images across unseen indoor scenes has not been systematically evaluated. Most existing computer vision perception models are developed for substantially high-resolution RGB or thermal imagery, and their generalization to native low-resolution thermal sensing across unseen indoor environments has not been systematically characterized.

Third, existing RGB--thermal image translation studies are commonly developed for a specific translation direction (e.g., RGB2T) or application setting, with relatively limited evaluation of bidirectional translation (e.g., T2RGB) under the same indoor conditions. Thermal-to-RGB and RGB-to-thermal translation are inherently different problems: the former requires inference of visible appearance that is absent from low-resolution thermal observations, whereas the latter requires estimation of thermal patterns that cannot be directly determined from visible appearance. Most studies have focused on the visual quality of the generated images in  RGB-to-thermal (RGB2T), but they lack research on the inverse thermal-to-RGB translation (T2RGB). A systematic evaluation of both directions on spatially aligned indoor data is needed.

In this work, we introduce a large-scale paired bedroom RGB--thermal dataset for low-resolution human perception and cross-modal analysis. The thermal images are captured at their native resolution of $80\times62$ pixels and paired with visible-light observations acquired from a camera mounted in the same enclosure with a fixed relative pose. The dataset covers 20 bedrooms and contains 90,576 thermal frames, 87,016 temporally matched RGB--thermal pairs, and 115,557 annotated human bounding boxes. The recordings include different person numbers and states, as well as acted and spontaneous sessions and scenes with heat-emitting interference. Rather than treating RGB--thermal translation as an isolated image-generation problem, we use the dataset to study both direct thermal perception and the transfer of task-relevant information across the two visual domains. We apply our dataset and deploy our models in a real-world sleep monitoring system, showing strong application value through field testing and commercial operation.

The main contributions of this work are summarized as follows:


\begin{itemize}[leftmargin=*]

\item We collect a large-scale paired RGB--thermal dataset for indoor human perception, covering 20 rooms and containing 90,576 thermal frames, 87,016 temporally matched RGB–thermal pairs, and 115,557 annotated human bounding boxes. 

\item We establish benchmarks for low-resolution thermal images, demonstrating the capability of native low-resolution thermal images for reliable occupancy detection, state recognition, and occupancy estimation under unseen bedroom conditions.



\item By comparing RGB-to-thermal generation methods and proposing a new thermal-to-RGB generation task, we establish cross-modal image generation benchmarks to evaluate cross-modal information transfer in unseen indoor environments.


\item We deploy the indoor thermal perception models developed using our dataset in a real-world sleep-monitoring system and validate their practical feasibility through long-term field testing and commercial operation.

\item  Our practical deployment reveals challenges and thus highlights future research directions, including environmental thermal interference, limited sensing range, and multi-person occlusion.



\end{itemize}


The codes, trained models and part of the dataset are available at https://github.com/kailaisun/RGBT and https://huggingface.co/skl24/RGBT. 


\section{Related Work}
\subsection{RGB--Thermal Datasets}

RGB--thermal datasets have been widely developed to support multimodal perception by leveraging the complementary information of visible and thermal modalities \cite{brenner2023rgb}. RGB images provide rich appearance information, while thermal images provide temperature information under varying illumination conditions.  In Table~\ref{tab:dataset_comparison}, early large-scale datasets mainly focused on outdoor perception tasks. For example, KAIST \cite{Hwang_2015_CVPR} and LLVIP \cite{jia2021llvip} were introduced for pedestrian detection in urban traffic scenarios, while M$^{3}$FD \cite{liu2022target} and MFNet \cite{ha2017mfnet} further extended RGB--thermal perception to multi-class object detection and semantic segmentation. More recently, large-scale RGB--thermal tracking datasets, such as LasHeR \cite{li2021lasher} and RGBDT500 \cite{zhu2026collaborating}, have advanced multimodal object tracking under diverse real-world conditions. 

Recently, RGB-thermal datasets have gradually been explored for indoor human perception. Indoor environments represent a major portion of daily human activities. Thermal-IM \cite{tang2023happened} investigated RGB--thermal human activity recognition in domestic environments, while the RGB-D-T dataset \cite{palmero2016multi} introduced aligned RGB-depth-thermal data for multimodal human segmentation in indoor scenarios. However, indoor (especially the bedroom) RGB--thermal datasets remain limited because of privacy issues \cite{laishram2025toward}. 

\subsection{Privacy-Preserving Thermal Human Perception}

Recent advances in human-centric sensing have focused on capturing fine-grained human information through visual and multimodal systems \cite{khan2025multimodal}. RGB cameras and depth sensors have demonstrated strong capabilities in occupancy detection, pose estimation, and activity understanding \cite{liu2024depth}. On the other hand, thermal sensing has emerged as a promising alternative for privacy-preserving indoor human perception \cite{dong2024privacy}. Unlike RGB imaging, thermal cameras capture infrared radiation patterns, enabling human perception while reducing exposure of identifiable information. Existing studies have explored thermal-based human perception for various applications, including occupancy detection, activity recognition, occupancy estimation, and indoor environmental control \cite{he2023smart,zhang2022room}.  Nevertheless, high-resolution visual sensing remains challenging for privacy-sensitive applications, motivating the exploration of compact and low-resolution thermal sensing solutions.
\subsection{System Deployment}


Although thermal sensors have demonstrated promising performance, translating these approaches into practical deployment requires considerations beyond model accuracy. Recent research has increasingly emphasized deployment-oriented evaluation of smart perception systems, including cost, computational efficiency, privacy-preserving data, and robustness to environmental changes \cite{liu2025towards}. Recent efforts have explored thermal sensors and AI edge inference for thermal perception systems \cite{farooq2022evaluation}. In particular, low-resolution thermal arrays combined with lightweight models offer cost-effective, privacy-aware perception while reducing hardware and computational requirements \cite{shubha2022real}. But the transition from laboratory evaluation to real-world deployment remains challenging, requiring systematic validation under diverse unseen scenarios and continuous improvement from deployment feedback.

\begin{table*}[t]
\centering
\caption{Comparison of representative RGB--thermal datasets.}
\label{tab:dataset_comparison}
\small
\setlength{\tabcolsep}{2pt}
\renewcommand{\arraystretch}{1.05}

\begin{tabular}{lcccccccc}
\toprule
Dataset
& Collection Environment
& \makecell{Thermal\\Resolution}
& Task
& BBox
& State
& Count
& \makecell{Thermal\\Interference}
& \makecell{RGB--Thermal\\Samples}
\\
\midrule

KAIST \cite{Hwang_2015_CVPR}
& Urban traffic
& $320\times256$
& Pedestrian detection
& \checkmark
& --
& --
& NR
& 95k
\\

LLVIP \cite{jia2021llvip}
& Urban traffic
& $1280\times720$
& Pedestrian detection
& \checkmark
& --
& --
& NR
& 15k
\\

M$^{3}$FD \cite{liu2022target}
& Urban traffic
& $640\times512$
& Object detection
& \checkmark
& --
& --
& NR
& 4k
\\

MFNet \cite{ha2017mfnet}
& Urban traffic
& $640\times480$
& Semantic segmentation
& --
& --
& --
& NR
& 1.5k
\\

DarkAct \cite{tan2026darkact}
& Indoor/outdoor
& NR
& Human action recognition
& --
& \checkmark
& --
& NR
& 12k clips
\\

RGBDT500 \cite{zhu2026collaborating}
& Indoor/outdoor
& $1920\times1080$
& Object tracking
& \checkmark
& --
& --
& NR
& 203.7k
\\

LasHeR \cite{li2021lasher}
& Indoor/outdoor
& NR
& Object tracking
& \checkmark
& --
& --
& NR
& 734.8k
\\

Tri-modal \cite{mogelmose2013tri}
& Indoor (1 room)
& $640\times480$
& Person re-identification
& --
& --
& --
& NR
& NR
\\

RGB-D-T Facial \cite{nikisins2014rgb}
& Indoor (1 room)
& $384\times288$
& Face recognition
& \checkmark
& --
& --
& NR
& 45.9k
\\

Thermal-IM \cite{tang2023happened}
& Indoor (2 rooms)
& $288\times384$
& Human activity recognition
& --
& \checkmark
& --
& NR
& 783 clips
\\

AAU RGB-D-T \cite{palmero2016multi}
& Indoor (3 scenes)
& $640\times480$
& Human segmentation
& --
& --
& --
& NR
& 11k
\\

\textbf{Indoor RGB-T (Ours)}
& \textbf{Indoor (20 rooms)}
& \textbf{$80\times62$}
& \textbf{\makecell{Privacy-preserving \\ human perception}}
& \checkmark
& \checkmark
& \checkmark
& \checkmark
& \textbf{87k frames}
\\

\bottomrule
\end{tabular}

\vspace{0.3em}

\parbox{0.98\linewidth}{\footnotesize
RGB--Thermal Samples denotes the number of paired RGB--thermal observations, frames, or clips reported by each dataset.
BBox indicates explicit bounding-box annotations.
State indicates human posture or action labels.
Count indicates frame-level human counting annotations.
Thermal Interference indicates whether challenging heat sources or temperature-related disturbances are explicitly considered during data collection.
NR denotes not reported.
}
\vspace{-1em}

\end{table*}

\section{Dataset}
\label{sec:dataset}

\subsection{Dataset Collection}
\label{subsec:data_collection}

\begin{figure}[tp]
    \centering
    \includegraphics[width=\linewidth]{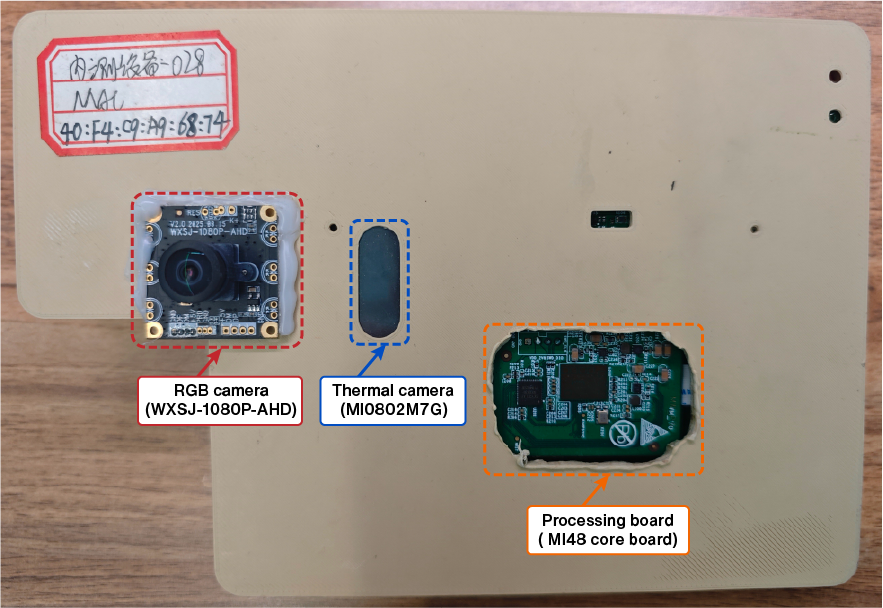}
    \vspace{-1em}
    \caption{\textbf{RGB--thermal data acquisition device.} The Meridian Innovation MI0802M7G thermal camera and a $640\times480$ visible-light camera are mounted in the same enclosure with a fixed relative pose.}
    \vspace{-1em}
    \label{fig:capture_device}
\end{figure}

The dataset was collected from June to October 2025 using a custom RGB--thermal acquisition device. The RGB and thermal cameras were integrated into the same enclosure with a fixed relative pose, enabling synchronized paired observations in Fig.~\ref{fig:capture_device}. Thermal frames were captured using a Meridian Innovation MI0802M7G thermal camera module from the SenXor MI0802 family, connected through an MI48 core board. The thermal camera has a native resolution of $80\times62$ pixels, with diagonal, horizontal, and vertical fields of view of $135^{\circ}$, $97^{\circ}$, and $73^{\circ}$, respectively. The camera module measures $9.0\times9.0\times5.5$~mm, and the thermal stream was recorded at approximately 7.3~Hz.

Each thermal image was retained at its native $80\times62$ resolution. No spatial upsampling or interpolation was applied to the thermal frames during preprocessing or visualization, and all bounding boxes were annotated directly in the native thermal image coordinate system. Each thermal image contains a 16-bit temperature field, which was additionally rendered as a pseudo-color thermal image for visualization.

The RGB camera recorded RGB images at a resolution of $640\times480$. The RGB and thermal streams were temporally synchronized during data collection. The fixed relative pose between the two cameras enabled spatial correspondence between the two modalities, resulting in paired RGB--thermal images for the following experiments.

All participants involved in the data collection were fully informed about the RGB–thermal recording procedure and voluntarily provided informed consent prior to participation. 

\subsection{Dataset Overview}
\label{sec:dataset-overview}

\begin{figure*}[tp]
\centering
\includegraphics[width=0.98\linewidth,height=0.98\textheight,keepaspectratio]{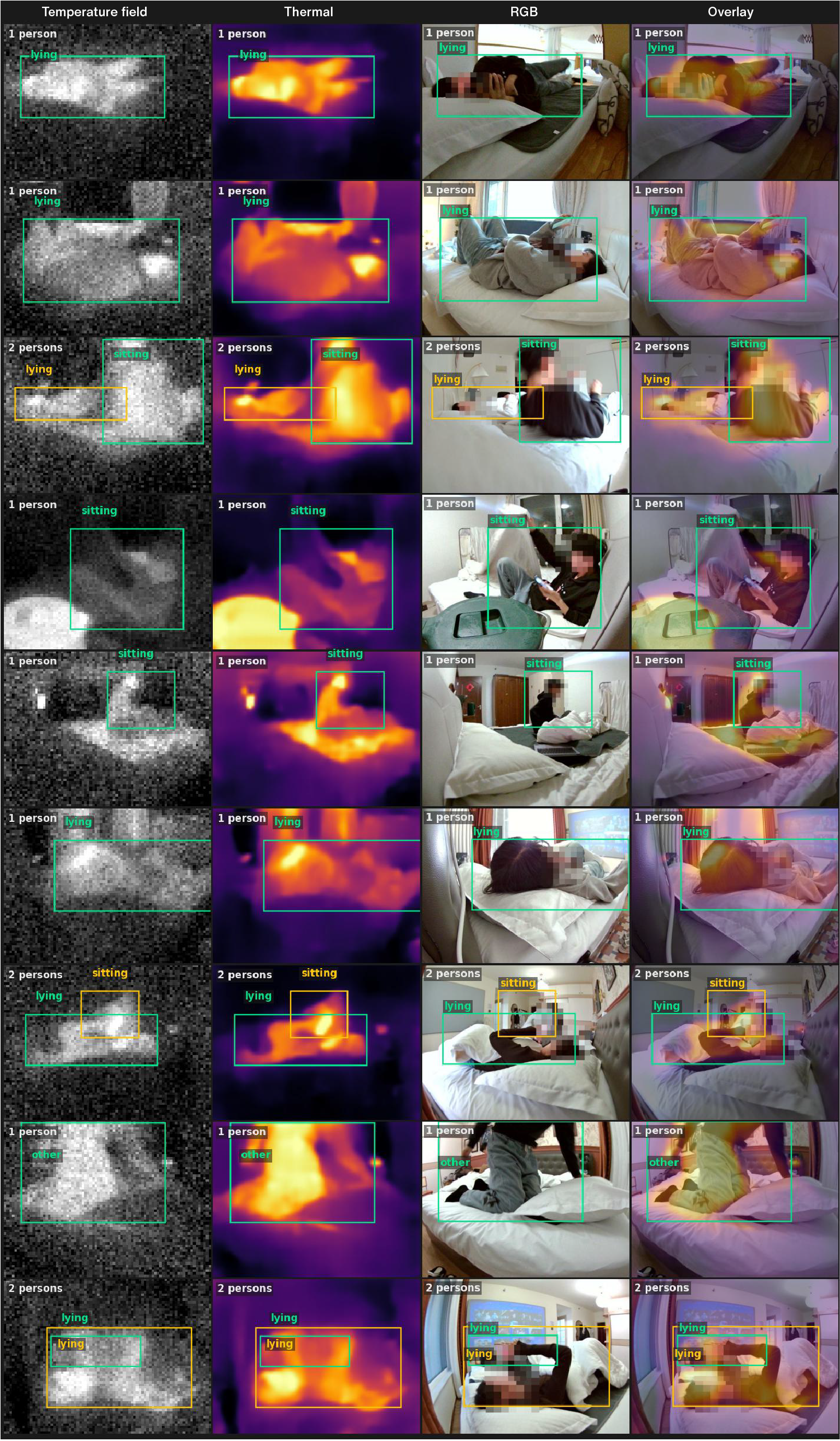}
\vspace{-1em}
\caption{Representative examples from the dataset, showing four modalities: temperature field, pseudo-color thermal image, RGB image, and RGB--thermal overlay. Bounding boxes are annotated in the native thermal image coordinate system and projected to the RGB image for reference.}
\vspace{-1em}
\label{fig:modality_grid}
\end{figure*}

Indoor RGB-T is a large-scale RGB--thermal dataset designed for human perception in indoor environments under native low-resolution thermal sensing. The dataset covers 20 indoor rooms and contains 86,751 annotated frames with 115,557 annotated human boxes. Among these annotated frames, 3,702 frames lack valid human-state annotations. After excluding these cases, 83,049 frames containing 110,781 occupant boxes have complete hierarchical annotations, including frame-level occupant counts, bounding boxes, and activity states. Figure~\ref{fig:modality_grid} presents representative examples from the dataset, including the temperature field, pseudo-color thermal image, RGB image, and RGB--thermal overlay.

The dataset covers diverse indoor conditions, including multiple-occupant scenarios, different human behaviors, recordings with external heat-source interference, and different thermal camera configurations across 20 rooms.

\subsection{Annotation}
\label{sec:annotation}

The dataset provides hierarchical annotations. The annotations are defined with respect to the thermal modality, while paired RGB images are used as auxiliary references to assist the annotation process.

At the frame level, each annotated frame is associated with a human-count annotation indicating the number of occupants present in the frame. This annotation is provided at the frame level, including challenging cases where occupant-level annotations are unavailable.

For frames with identifiable occupants, occupant-level annotations are further provided. Each occupant is annotated with a bounding box directly in the native thermal image coordinate system. Each bounding box is associated with hierarchical semantic labels. First, a location label indicates whether the occupant is on the bed or off the bed. For on-bed occupants, an additional state label is provided, including \textit{lying}, \textit{sitting}, and \textit{other}. Combining the location and state annotations results in four mutually exclusive occupant states: \textit{lying}, \textit{sitting}, \textit{other}, and \textit{off-bed}.

A total of 3,702 frames lack valid occupant-state annotations. Among them, 668 frames contain no occupant, 1,192 frames contain three or more occupants and are annotated only with frame-level counts, 1,519 frames contain thermal observations where occupants cannot be reliably localized, and 323 frames contain occupant bounding boxes but lack sufficient information to determine one of the four occupant states.

\begin{figure*}[tp]
  \centering
  \includegraphics[width=0.9\textwidth]{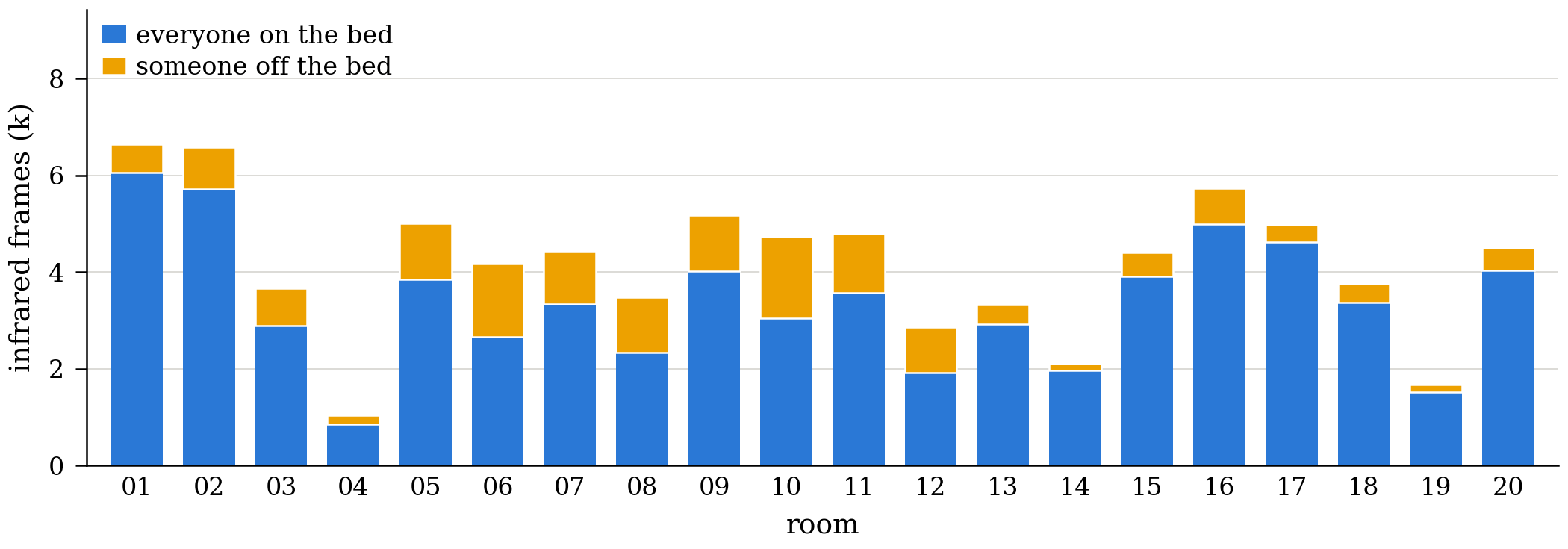}
  \vspace{-1em}
  \caption{Room-level occupancy composition of the dataset. Each bar represents one room and shows the distribution of annotated frames with all occupants on the bed or at least one occupant off the bed.}
  \vspace{-1em}
  \label{fig:composition}
\end{figure*}

\begin{table*}[ht!]
\caption{Dataset statistics and acquisition conditions. Rooms are numbered \texttt{room01}--\texttt{room20}; the mapping to recording sessions and venues is provided in the dataset room index.}
\vspace{-0.5em}
\begin{center}
\begin{tabular*}{\textwidth}{@{\extracolsep{\fill}}l rrrr rr rr c@{}}
\toprule
\multirow{2}{*}{Room} & \multicolumn{4}{c}{Frames} & \multicolumn{2}{c}{Density}
 & \multicolumn{2}{c}{Interference} & \multirow{2}{*}{Cams} \\
\cmidrule(lr){2-5}\cmidrule(lr){6-7}\cmidrule(lr){8-9}
 & Thermal & Annot. & RGB & Pairs & Boxes & Occ./fr. & Without & With & \\
\midrule
\texttt{room01} &  6,933 &  6,930 & 4,867 & 6,907 &   8,991 & 1.30 &    -- &    -- & 2 \\
\texttt{room02} &  6,746 &  6,746 & 4,615 & 6,746 &   8,838 & 1.31 &    -- &    -- & 2 \\
\texttt{room03} &  4,278 &  3,938 & 1,515 & 3,936 &   5,263 & 1.34 &    -- &    -- & 1 \\
\texttt{room04} &  1,056 &  1,056 & 1,053 & 1,053 &   1,233 & 1.17 &   516 &   540 & 1 \\
\texttt{room05} &  5,152 &  5,135 & 5,091 & 5,096 &   6,246 & 1.22 & 1,357 & 3,795 & 1 \\
\texttt{room06} &  4,488 &  4,443 & 4,488 & 4,487 &   5,338 & 1.20 &   929 & 3,559 & 1 \\
\texttt{room07} &  4,612 &  4,542 & 4,594 & 4,594 &   5,700 & 1.25 & 1,204 & 3,408 & 1 \\
\texttt{room08} &  3,617 &  3,572 & 3,617 & 3,617 &   4,302 & 1.20 &   825 & 2,792 & 1 \\
\texttt{room09} &  5,372 &  5,341 & 5,372 & 5,372 &   6,918 & 1.30 & 1,771 & 3,601 & 1 \\
\texttt{room10} &  5,009 &  4,974 & 5,004 & 5,003 &   6,622 & 1.33 &   934 & 4,075 & 1 \\
\texttt{room11} &  4,978 &  4,878 & 4,978 & 4,978 &   5,974 & 1.22 & 1,083 & 3,895 & 1 \\
\texttt{room12} &  3,324 &  3,199 & 3,324 & 3,324 &   4,485 & 1.40 &   632 & 2,692 & 1 \\
\texttt{room13} &  4,304 &  3,491 & 1,520 & 3,462 &   4,920 & 1.41 &    -- &    -- & 2 \\
\texttt{room14} &  3,416 &  2,216 & 1,264 & 2,209 &   2,715 & 1.23 &    -- &    -- & 1 \\
\texttt{room15} &  4,890 &  4,607 & 1,503 & 4,607 &   6,844 & 1.49 &    -- &    -- & 2 \\
\texttt{room16} &  6,326 &  5,862 & 1,418 & 5,854 &   8,068 & 1.38 &    -- &    -- & 2 \\
\texttt{room17} &  5,610 &  5,426 & 1,692 & 5,426 &   8,216 & 1.51 &    -- &    -- & 2 \\
\texttt{room18} &  3,874 &  3,835 & 2,833 & 3,818 &   5,781 & 1.51 &    -- &    -- & 2 \\
\texttt{room19} &  1,777 &  1,747 & 1,617 & 1,719 &   2,373 & 1.36 &    -- &    -- & 1 \\
\texttt{room20} &  4,814 &  4,813 & 3,423 & 4,808 &   6,730 & 1.40 &    -- &    -- & 2 \\
\midrule
Total & \textbf{90,576} & \textbf{86,751} & \textbf{63,788} & \textbf{87,016} & \textbf{115,557} & 1.33 & \textbf{9,251} & \textbf{28,357} & \textbf{28} \\
\bottomrule
\end{tabular*}

\vspace{0.3em}
\parbox{0.95\linewidth}{\footnotesize Thermal denotes the number of recorded thermal frames. Annot. denotes the number of annotated frames. RGB denotes the number of recorded RGB images. Pairs denotes thermal frames with a temporally matched RGB counterpart ($\leq$500\,ms). Boxes denote the number of annotated occupant bounding boxes. Occ./fr. denotes the average number of occupants per annotated frame. Interference indicates frames recorded without and with external heat-source interference. Cams denotes the number of thermal cameras used for recording each room.}
\vspace{-1em}
\label{dataset}
\end{center}
\end{table*}

\begin{figure*}[tp]
  \centering
  \includegraphics[width=0.9\textwidth]{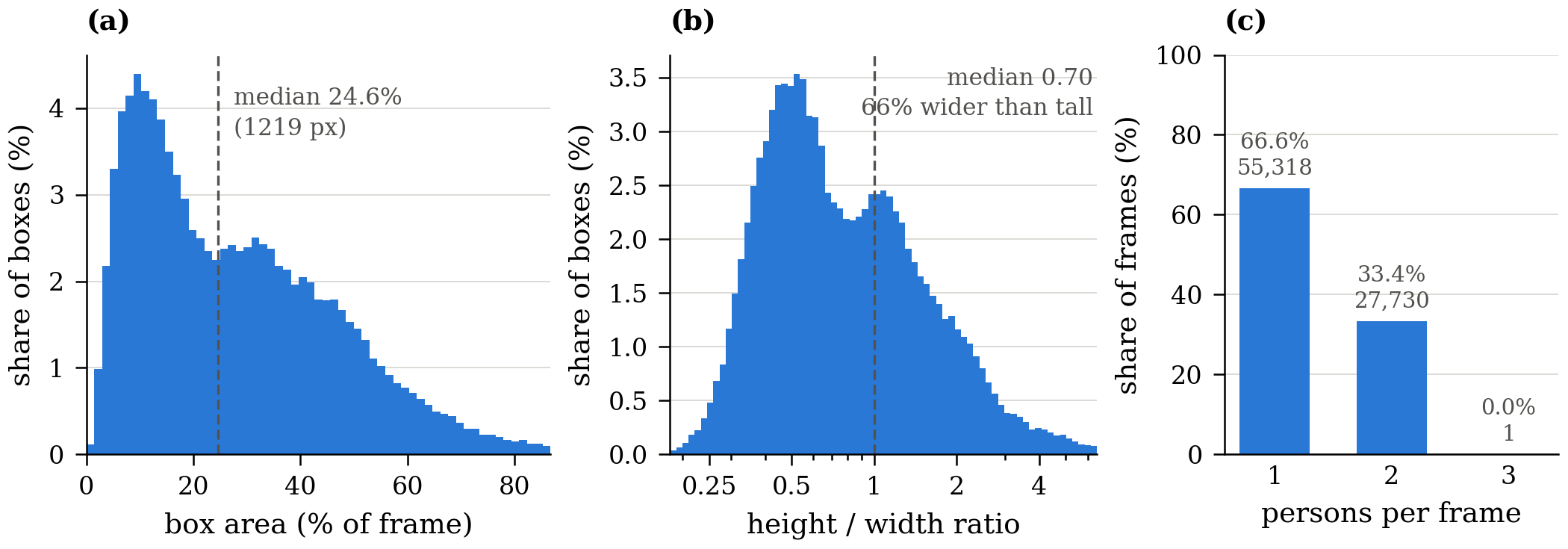}
  \vspace{-1em}
  \caption{Geometry statistics of occupant bounding boxes. The distributions are computed over 110,781 boxes from 83,049 frames. \textbf{(a)} Box area distribution. \textbf{(b)} Height-to-width ratio distribution. \textbf{(c)} Occupant number distribution per frame.}
  \vspace{-1em}
  \label{fig:box-geometry}
\end{figure*}

\begin{figure*}[tp]
  \centering
  \includegraphics[width=0.9\textwidth]{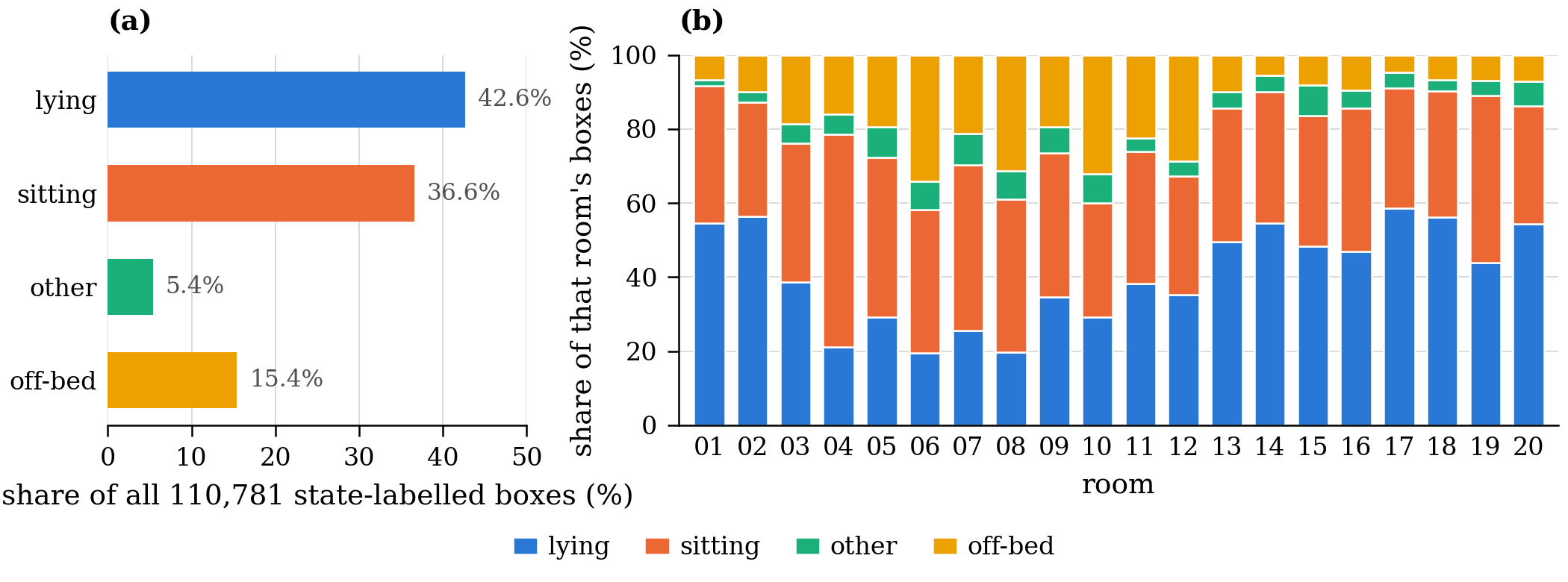}
  \vspace{-1em}
 \caption{Occupant-state distribution over 110,781 boxes from 83,049 frames. \textbf{(a)} Overall distribution of the four occupant states. \textbf{(b)} Room-level state composition normalized by the number of annotated boxes in each room.}
 \vspace{-1em}
  \label{fig:posture}
\end{figure*}

\subsection{Dataset Characteristics}
\label{sec:dataset-stats}

The dataset characteristics are analyzed from three perspectives: room-level composition, occupant-box geometry, and occupant-state distribution. Figure~\ref{fig:composition} and Table~\ref{dataset} summarize the room-level statistics of the dataset. The dataset contains annotated frames from 20 indoor bedrooms, with the number of annotated frames ranging from 1,056 to 6,930 per room. Table~\ref{dataset} also summarizes the acquisition conditions, including spontaneous recordings with and without external heat-source interference and different numbers of thermal cameras across rooms. Figure~\ref{fig:composition} further presents the occupancy composition across rooms. Overall, 67,606 frames (81.4\%) contain occupants entirely on the bed, whereas 15,443 frames (18.6\%) include at least one occupant off the bed. The proportion of off-bed cases varies across rooms, ranging from 6.7\% to 36.3\%.

Figure~\ref{fig:box-geometry} summarizes the geometric characteristics of the occupant bounding boxes. Figure~\ref{fig:box-geometry}(a) shows the distribution of box areas relative to the $80\times62$ thermal frame. The median box area is 24.6\% of the frame, corresponding to 1,219 pixels. The box areas exhibit a wide distribution, with 25\% of the boxes occupying more than 40.6\% of the frame and only 5\% occupying less than 5.2\%. Figure~\ref{fig:box-geometry}(b) presents the height-to-width ratio distribution of the bounding boxes. The median ratio is 0.70, and 65.7\% of the boxes have ratios below 1, indicating that most occupant boxes are wider than they are tall. The remaining boxes have ratios above 1, with the overall distribution ranging from 0.18 to 6.5. Figure~\ref{fig:box-geometry}(c) presents the occupant number distribution per frame. Among the 83,049 frames, 55,318 frames (66.6\%) contain one occupant, 27,730 frames (33.4\%) contain two occupants, and only one frame contains three occupants.

Figure~\ref{fig:posture} summarizes the distribution of occupant states over the 110,781 boxes with valid state annotations. The dataset contains four mutually exclusive states: \textit{lying}, \textit{sitting}, \textit{other}, and \textit{off-bed}. Overall, \textit{lying} and \textit{sitting} constitute the majority of annotated boxes, accounting for 79.2\% in total, whereas \textit{other} represents 5.4\% of the boxes. The state distribution varies across rooms, with the proportion of \textit{lying} ranging from 19.5\% to 58.6\% and the proportion of \textit{off-bed} ranging from 4.7\% to 34.1\%. Figure~\ref{fig:posture}(b) further presents the room-level state composition, where each room is normalized independently based on its annotated boxes.



\section{Benchmark Methodology}

\subsection{Problem Definition}

We formulate indoor thermal perception as a multi-task benchmark for privacy-preserving occupant understanding using native low-resolution thermal sensing. The benchmark comprises two components: the thermal perception benchmark and cross-modal RGB--thermal translation.

For the thermal perception benchmark, let $X_T$ denote a native $80\times62$ thermal image. We consider three tasks: occupancy detection, occupant state recognition, and occupancy estimation. occupancy detection localizes occupants from $X_T$. Occupant state recognition distinguishes four states: \textit{lying}, \textit{sitting}, \textit{off-bed}, and \textit{other}. Occupancy estimation performs image-level classification between one-occupant and two-occupant cases.

For cross-modal RGB--thermal translation, let $X_R$ denote the paired RGB image. We evaluate both RGB-to-thermal and thermal-to-RGB directions. RGB-to-thermal translation estimates thermal patterns and temperature-related characteristics from $X_R$, while thermal-to-RGB synthesis generates visible representations from $X_T$. Together, these tasks evaluate cross-modal information correspondence and recoverability.

\subsection{Evaluation Protocol}

To ensure fair comparison across different tasks, we adopt a room-disjoint evaluation protocol for the thermal perception benchmark. The dataset is partitioned at the room level to avoid information leakage from shared environments and temporally adjacent observations. Rooms 07, 13, and 20 are used for validation, while Rooms 03, 10, and 18 are reserved for testing, and the remaining rooms are used for training. 

For occupant perception tasks, each thermal observation $X_T$ is associated with task-specific annotations. For occupancy detection, the target is defined as:$
Y_D=\{(b_i,c_i)\}_{i=1}^{N},$
where $b_i$ denotes the bounding box of the $i$-th occupant and $c_i$ represents the corresponding occupant state label. 

For occupancy estimation, the task is formulated as binary classification:
$
Y_O\in\{1,2\},
$
where the label indicates one-occupant or two-occupant cases. This is because a 1-2 occupancy estimate covers the primary occupancy configurations in bedroom. The accuracy and balanced accuracy are reported:
$
\mathrm{BAcc}=
\frac{1}{2}
\left(
\frac{TP}{TP+FN}
+
\frac{TN}{TN+FP}
\right).
$

For cross-modal translation tasks, paired RGB--thermal observations $(X_R,X_T)$ are used. RGB-to-thermal translation aims to reconstruct the thermal observation:
\[
\hat{X}_T=G_{RT}(X_R),
\]
while thermal-to-RGB synthesis generates a visible representation:
\[
\hat{X}_R=G_{TR}(X_T).
\]

For RGB-to-thermal translation, temperature reconstruction accuracy is evaluated using MAE, RMSE, and $R^2$. For image translation quality, metrics are reported, including PSNR, SSIM, MS-SSIM, LPIPS, $\Delta E$, and FID.





\vspace{-1em}
\section{Experiments and Discussion}
\label{sec:exp setting}

\subsection{Experimental Setup
}

Experiments were conducted on a 64-bit Ubuntu 22.04 platform equipped with an AMD EPYC 9754 128-Core Processor, 1.5 TB RAM, and 10 NVIDIA L40S GPUs (48 GB memory each), running CUDA 12.0. The programming environment was Python 3.10 and PyTorch 1.13.

\subsection{Thermal Perception Benchmark}

For all thermal perception experiments, only $X_T$ is used as visual input, without paired RGB information. A common room-disjoint split is adopted to avoid information leakage across indoor environments. Inference efficiency is measured on a single GPU with batch size 1.

\subsubsection{Thermal occupancy detection}

For occupancy detection and state recognition, we benchmark six representative detectors spanning two-stage, one-stage, and Transformer-based architectures: Faster R-CNN \cite{ren2016faster}, RTMDet \cite{lyu2022rtmdet}, DINO \cite{zhang2023dino}, YOLO11 \cite{khanam2024yolov11}, RT-DETR \cite{zhao2024detrs}, and YOLO26 \cite{sapkota2025yolo26}. 
Large model-based detection methods (e.g., large computer vision models) are not feasible for practical deployment locally. All six models are initialized with their official COCO-pretrained weights and fine-tuned on the training rooms. Each prediction consists of an occupant bounding box and one of four state labels: \textit{lying}, \textit{sitting}, \textit{other}, or \textit{off-bed}.

Table~\ref{tab:thermal_detection} summarizes the detection performance on the room-disjoint test set. All evaluated models achieve reliable occupancy detection on previously unseen indoor environments, with $\mathrm{mAP}_{50}$ values ranging from 0.681 to 0.751. YOLO26 achieves the best overall detection performance, obtaining an $\mathrm{mAP}_{50}$ of 0.751 and an $\mathrm{mAP}_{50:95}$ of 0.480, followed by RTMDet with $\mathrm{mAP}_{50}=0.742$ and $\mathrm{mAP}_{50:95}=0.452$. YOLO26 also achieves the highest precision (0.778) and macro F1-score (0.723), whereas YOLO11 obtains the highest recall (0.681) among all evaluated models. The remaining detectors, including Faster R-CNN, DINO, and RT-DETR, also perform competitively, with $\mathrm{mAP}_{50}$ above 0.68.

The inference efficiency results further demonstrate the feasibility of real-time occupancy detection using low-resolution thermal inputs. YOLO11 provides the highest throughput, achieving 187.3 FPS with a latency of 5.339 ms per frame, followed by YOLO26 with 171.6 FPS and 5.828 ms latency. RTMDet and RT-DETR achieve comparable inference speeds of 83.0 and 85.8 FPS, respectively, while Faster R-CNN reaches 76.7 FPS. DINO has the highest computational cost among the evaluated models, with a latency of 24.378 ms per frame and an inference speed of 41.0 FPS.

\begin{table*}[t]
\centering
\caption{Thermal occupancy detection performance on the room-disjoint test set.}
\label{tab:thermal_detection}
\small
\setlength{\tabcolsep}{4.5pt}
\renewcommand{\arraystretch}{1.10}
\begin{tabular}{lccccccc}
\toprule
Method
& $\mathrm{mAP}_{50}$
& $\mathrm{mAP}_{50:95}$
& Precision
& Recall
& Macro F1
& Latency (ms)
& FPS \\
\midrule

Faster R-CNN \cite{ren2016faster}
& 0.723
& 0.412
& 0.751
& 0.649
& 0.640
& 13.036
& 76.7 \\

RTMDet \cite{lyu2022rtmdet}
& 0.742
& 0.452
& 0.751
& 0.666
& 0.699
& 12.049
& 83.0 \\

DINO \cite{zhang2023dino}
& 0.681
& 0.438
& 0.686
& 0.613
& 0.630
& 24.378
& 41.0 \\

YOLO11 \cite{khanam2024yolov11}
& 0.705
& 0.427
& 0.704
& \textbf{0.681}
& 0.692
& \textbf{5.339}
& \textbf{187.3} \\

RT-DETR \cite{zhao2024detrs}
& 0.692
& 0.393
& 0.733
& 0.633
& 0.668
& 11.659
& 85.8 \\

YOLO26 \cite{sapkota2025yolo26}
& \textbf{0.751}
& \textbf{0.480}
& \textbf{0.778}
& 0.680
& \textbf{0.723}
& 5.828
& 171.6 \\

\bottomrule
\end{tabular}
\vspace{-1em}
\end{table*}

\subsubsection{Thermal Occupant State Recognition}

Using the same detector outputs, we further analyze the class-wise recognition performance of the four states. 

Table~\ref{tab:thermal_state} summarizes the class-wise recognition performance of different detector architectures on the test set. Across all evaluated models, the four occupant states exhibit distinct recognition difficulty. The \textit{lying} and \textit{sitting} states achieve the strongest performance, with $\mathrm{mAP}_{50}$ ranges of 0.795--0.897 and 0.764--0.840, respectively. The \textit{off-bed} state also shows relatively stable performance across models, with $\mathrm{mAP}_{50}$ ranging from 0.680 to 0.773. In contrast, the \textit{other} state remains the most challenging category, with $\mathrm{mAP}_{50}$ values between 0.379 and 0.562. This class contains more diverse occupant appearances and less consistent thermal patterns, resulting in larger performance variations among different detectors.

The comparison across architectures reveals complementary strengths among the evaluated models. YOLO26 achieves the strongest overall state recognition performance, obtaining the highest $\mathrm{mAP}_{50}$ for \textit{sitting} (0.840) and \textit{other} (0.562), together with the highest corresponding F1-scores of 0.779 and 0.589. Faster R-CNN achieves the best performance for \textit{lying} and \textit{off-bed}, reaching $\mathrm{mAP}_{50}$ values of 0.897 and 0.773, respectively. These results indicate that different detector architectures exhibit different advantages across occupant states.

\begin{figure*}[tp]
  \centering
  \includegraphics[width=0.8\textwidth]{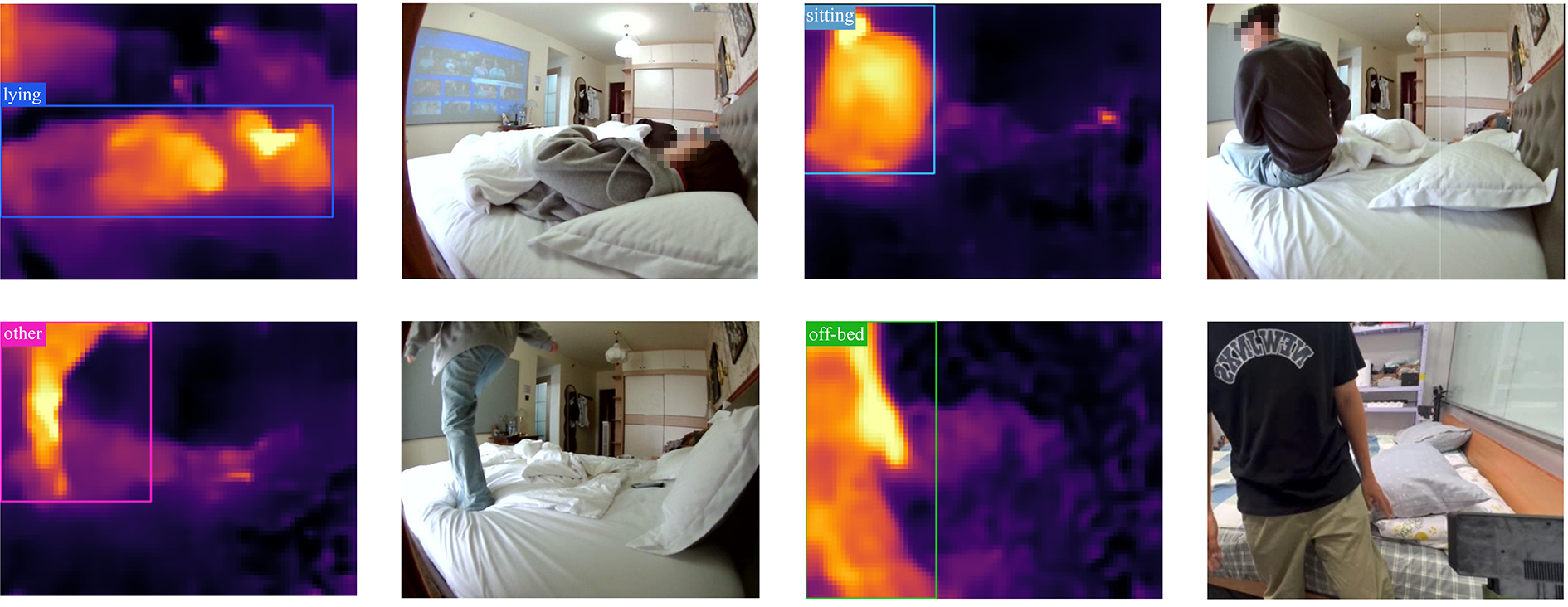}
  \vspace{-1em}
  \caption{Qualitative examples of thermal occupancy detection with state recognition for four occupant states: lying, sitting, other, and off-bed.}
  \vspace{-1em}
  \label{fig:results}
\end{figure*}

Figure~\ref{fig:results} presents qualitative examples of thermal occupancy detection with state recognition. The predicted bounding boxes successfully localize occupants and identify their corresponding states, including \textit{lying}, \textit{sitting}, \textit{other}, and \textit{off-bed}.

\begin{table}[t]
\centering
\caption{Class-wise thermal occupant state recognition performance on the room-disjoint test set.}
\label{tab:thermal_state}
\small
\setlength{\tabcolsep}{1pt}
\renewcommand{\arraystretch}{1.08}
\begin{tabular}{llccccc}
\toprule
Method
& State
& $\mathrm{mAP}_{50}$
& $\mathrm{mAP}_{50:95}$
& Precision
& Recall
& F1-score \\
\midrule

\multirow{4}{*}{\makecell[c]{Faster \\ R-CNN}\cite{ren2016faster}}
& Lying
& \textbf{0.897}
& 0.482
& 0.676
& \textbf{0.905}
& 0.774 \\
& Sitting
& 0.834
& 0.486
& 0.714
& \textbf{0.839}
& 0.771 \\
& Other
& 0.387
& 0.264
& \textbf{0.828}
& 0.174
& 0.288 \\
& Off-bed
& \textbf{0.773}
& 0.415
& 0.786
& 0.679
& \textbf{0.728} \\

\midrule

\multirow{4}{*}{RTMDet \cite{lyu2022rtmdet}}
& Lying
& 0.860
& 0.487
& 0.872
& 0.715
& 0.786 \\
& Sitting
& 0.794
& 0.518
& 0.721
& 0.758
& 0.739 \\
& Other
& 0.544
& 0.360
& 0.715
& 0.457
& 0.558 \\
& Off-bed
& 0.771
& \textbf{0.442}
& 0.693
& 0.734
& 0.713 \\

\midrule

\multirow{4}{*}{DINO \cite{zhang2023dino}}
& Lying
& 0.840
& \textbf{0.533}
& 0.673
& 0.846
& 0.750 \\
& Sitting
& 0.764
& 0.531
& 0.678
& 0.752
& 0.713 \\
& Other
& 0.379
& 0.273
& 0.547
& 0.296
& 0.384 \\
& Off-bed
& 0.739
& 0.412
& \textbf{0.848}
& 0.558
& 0.673 \\

\midrule

\multirow{4}{*}{YOLO11 \cite{khanam2024yolov11}}
& Lying
& 0.852
& 0.509
& 0.867
& 0.782
& \textbf{0.822} \\
& Sitting
& 0.815
& 0.510
& 0.754
& 0.790
& 0.772 \\
& Other
& 0.452
& 0.294
& 0.498
& 0.468
& 0.482 \\
& Off-bed
& 0.702
& 0.398
& 0.698
& 0.685
& 0.691 \\

\midrule

\multirow{4}{*}{RT-DETR \cite{zhao2024detrs}}
& Lying
& 0.795
& 0.449
& \textbf{0.884}
& 0.680
& 0.769 \\
& Sitting
& 0.781
& 0.451
& 0.748
& 0.719
& 0.733 \\
& Other
& 0.511
& 0.341
& 0.690
& 0.395
& 0.502 \\
& Off-bed
& 0.680
& 0.330
& 0.611
& \textbf{0.738}
& 0.669 \\

\midrule

\multirow{4}{*}{YOLO26 \cite{sapkota2025yolo26}}
& Lying
& 0.856
& 0.530
& 0.826
& 0.802
& 0.814 \\
& Sitting
& \textbf{0.840}
& \textbf{0.563}
& \textbf{0.803}
& 0.755
& \textbf{0.779} \\
& Other
& \textbf{0.562}
& \textbf{0.387}
& 0.634
& \textbf{0.549}
& \textbf{0.589} \\
& Off-bed
& 0.748
& 0.440
& 0.847
& 0.615
& 0.713 \\

\bottomrule
\end{tabular}
\vspace{-1em}
\end{table}

\subsubsection{Occupancy Estimation}

We compare seven pretrained backbones, including ResNet18, ConvNeXt-T, ConvNeXt-S, Swin-T, Swin-S, ViT-B/16, and DINOv2 ViT-S. The aspect ratio of each thermal image is preserved before padding to $224\times224$. Models are trained for 30 epochs using weighted cross-entropy, AdamW, and cosine learning-rate scheduling. The default learning rate is set to $3\times10^{-4}$, while ViT-B/16 uses $1\times10^{-4}$. The checkpoint with the highest validation macro F1 is selected for testing. 
The test set contains 63.3\% one-occupant and 36.7\% two-occupant frames; therefore, ROC-AUC is additionally reported for the less frequent two-occupant class. 

Table~\ref{tab:occupancy_estimation} summarizes the occupancy estimation performance on the test set with two classes. Among the evaluated backbones, Swin-S achieves the best overall performance, obtaining an accuracy of 0.847, balanced accuracy of 0.843, and macro F1 of 0.838. It also reliably discriminates the two-occupant class, achieving a recall of 0.827, an F1 score of 0.799, and a ROC-AUC of 0.924. The one-occupant and two-occupant F1 scores are 0.876 and 0.799, respectively, indicating that the model maintains relatively balanced performance between the dominant and minority occupancy levels.

The results further show that convolutional and hierarchical Transformer backbones perform consistently well on the low-resolution thermal inputs. ConvNeXt-T achieves a macro F1 of 0.829, which is close to Swin-S, while Swin-T obtains a macro F1 of 0.819. In particular, ConvNeXt-T provides a more favorable accuracy-efficiency trade-off, achieving only a small decrease in macro F1 compared with Swin-S but reducing the median latency from 15.74 ms to 5.03 ms. These results suggest that lightweight architectures can retain strong occupancy estimation capability even when the input resolution is limited to $80\times62$.

Increasing model capacity does not consistently improve generalization in this task. ConvNeXt-S achieves the highest one-occupant recall (0.893) and two-occupant precision (0.804), but its two-occupant recall decreases to 0.754, resulting in a macro F1 of 0.828, slightly lower than ConvNeXt-T. This indicates that larger backbones do not necessarily provide better performance in this scenario. 
DINOv2 ViT-S achieves a lower overall performance, with an accuracy of 0.702 and ROC-AUC of 0.745, suggesting that generic visual representations may not directly transfer effectively to extremely low-resolution thermal inputs.

Overall, the occupancy estimation results demonstrate that the proposed dataset supports reliable occupancy information extraction directly from $80\times62$ thermal images without RGB information under unseen room conditions.


\begin{table*}[t]
\centering
\caption{Occupancy estimation performance on the room-disjoint test set. 
1-Occ. and 2-Occ. denote the one-occupant and two-occupant classes, respectively. 
Latency is reported in milliseconds.}
\label{tab:occupancy_estimation}
\small
\setlength{\tabcolsep}{3pt}
\renewcommand{\arraystretch}{1.1}

\begin{tabular}{@{}lccccccccccc@{}}
\toprule
Method
& Acc
& BAcc
& Macro F1
& 1Occ. P
& 1Occ. R
& 1Occ. F1
& 2Occ. P
& 2Occ. R
& 2Occ. F1
& ROC-AUC
& Latency \\
\midrule

ConvNeXt-T \cite{liu2022convnet}
& 0.841
& 0.829
& 0.829
& 0.873
& 0.877
& 0.875
& 0.786
& 0.780
& 0.783
& 0.909
& 5.03 \\

ConvNeXt-S \cite{liu2022convnet}
& 0.842
& 0.824
& 0.828
& 0.862
& \textbf{0.893}
& \textbf{0.877}
& \textbf{0.804}
& 0.754
& 0.778
& 0.908
& 11.46 \\

Swin-T \cite{liu2021swinv2}
& 0.830
& 0.823
& 0.819
& 0.880
& 0.846
& 0.863
& 0.752
& 0.801
& 0.775
& 0.909
& 6.54 \\

ResNet18 \cite{he2016deep}
& 0.817
& 0.808
& 0.805
& 0.866
& 0.839
& 0.853
& 0.738
& 0.777
& 0.757
& 0.892
& \textbf{2.79} \\

ViT-B/16 \cite{dosovitskiy2020image}
& 0.820
& 0.805
& 0.806
& 0.855
& 0.861
& 0.858
& 0.758
& 0.750
& 0.754
& 0.895
& 3.40 \\

DINOv2 ViT-S \cite{oquab2023dinov2}
& 0.702
& 0.679
& 0.679
& 0.764
& 0.764
& 0.764
& 0.594
& 0.595
& 0.595
& 0.745
& 3.84 \\

Swin-S \cite{liu2021swinv2}
& \textbf{0.847}
& \textbf{0.843}
& \textbf{0.838}
& \textbf{0.895}
& 0.858
& 0.876
& 0.772
& \textbf{0.827}
& \textbf{0.799}
& \textbf{0.924}
& 15.74 \\

\bottomrule
\end{tabular}
\end{table*}

\subsection{Cross-modal RGB--Thermal Translation}

We investigate cross-modal translation between thermal and RGB modalities to evaluate whether complementary visual representations can be inferred from paired observations. The experiments are conducted on RGB--thermal paired observations collected from the same rooms. For both translation directions, the thermal modality remains at its $80\times62$ resolution, and the original spatial correspondence between RGB and thermal modalities is preserved.
\subsubsection{RGB-to-Thermal Translation}
\label{subsec:rgb2t}

Given an RGB image, the RGB-to-thermal translation task aims to predict the corresponding thermal representation of the same room. Experiments are conducted on paired observations from \texttt{room04}--\texttt{room12} using a 7:2 room training/test split with a fixed random seed of 42. The raw thermal measurements are converted from 16-bit values into temperature fields in degrees Celsius and used as the ground truth. We compare four representative translation methods, including U-Net, Pix2Pix, ThermalGAN, and Palette DDPM. The generated thermal outputs are evaluated using temperature-level and image-level metrics. Temperature accuracy is measured using MAE, RMSE, and pooled $R^2$, while image-level similarity is evaluated using PSNR, SSIM, and LPIPS. For image-level evaluation, predicted temperature fields are rendered using the same fixed Inferno colormap as the ground-truth thermal fields before metric computation.

\begin{table*}[t]
\centering
\caption{RGB-to-thermal translation performance on the test set.
Lower MAE, RMSE, and LPIPS are better, whereas higher $R^2$, PSNR, and
SSIM are better.}
\label{tab:rgb2t}
\small
\setlength{\tabcolsep}{8pt}
\renewcommand{\arraystretch}{1.10}
\begin{tabular}{lcccccc}
\toprule
Method &
MAE ($^{\circ}$C)$\downarrow$ &
RMSE ($^{\circ}$C)$\downarrow$ &
$R^2$$\uparrow$ &
PSNR$\uparrow$ &
SSIM$\uparrow$ &
LPIPS$\downarrow$ \\
\midrule

Pix2Pix \cite{isola2017image}
& 0.753
& 1.000
& 0.896
& 25.519
& 0.572
& 0.068 \\

ThermalGAN \cite{xiao2025thermalgen}
& 0.726
& 0.957
& 0.905
& 25.853
& 0.582
& \textbf{0.063} \\

Palette DDPM \cite{saharia2022palette} 
& 2.236
& 2.892
& 0.134
& 18.752
& 0.466
& 0.166 \\

U-Net \cite{ronneberger2015u}
& \textbf{0.654}
& \textbf{0.860}
& \textbf{0.923}
& \textbf{26.754}
& \textbf{0.616}
& 0.176 \\

\bottomrule
\end{tabular}
\end{table*}

Table~\ref{tab:rgb2t} summarizes the quantitative results of RGB-to-thermal translation. For temperature-level reconstruction, U-Net achieves the best temperature reconstruction performance among the evaluated methods, obtaining an MAE of $0.654\,^{\circ}\mathrm{C}$, an RMSE of $0.860\,^{\circ}\mathrm{C}$, and an $R^2$ of 0.923. ThermalGAN-style and Pix2Pix achieve comparable temperature estimation performance, with MAEs of $0.726\,^{\circ}\mathrm{C}$ and $0.753\,^{\circ}\mathrm{C}$, respectively. In contrast, Palette DDPM produces substantially larger temperature deviations, achieving an MAE of $2.236\,^{\circ}\mathrm{C}$ and an $R^2$ of only 0.134. 

For image-level similarity, different metrics favor different methods. U-Net achieves the highest PSNR of 26.754~dB, indicating the strongest pixel-level correspondence with the ground-truth thermal images. ThermalGAN-style achieves the lowest LPIPS of 0.063, followed closely by Pix2Pix with 0.068, while U-Net obtains an SSIM of 0.616. These results indicate that temperature reconstruction accuracy and perceptual similarity are not fully aligned. U-Net provides the most accurate recovery of the underlying thermal values and the highest PSNR, whereas the adversarial models achieve stronger perceptual similarity as reflected by their lower LPIPS values.

\begin{figure*}[tp]
  \centering
  \includegraphics[width=\textwidth]{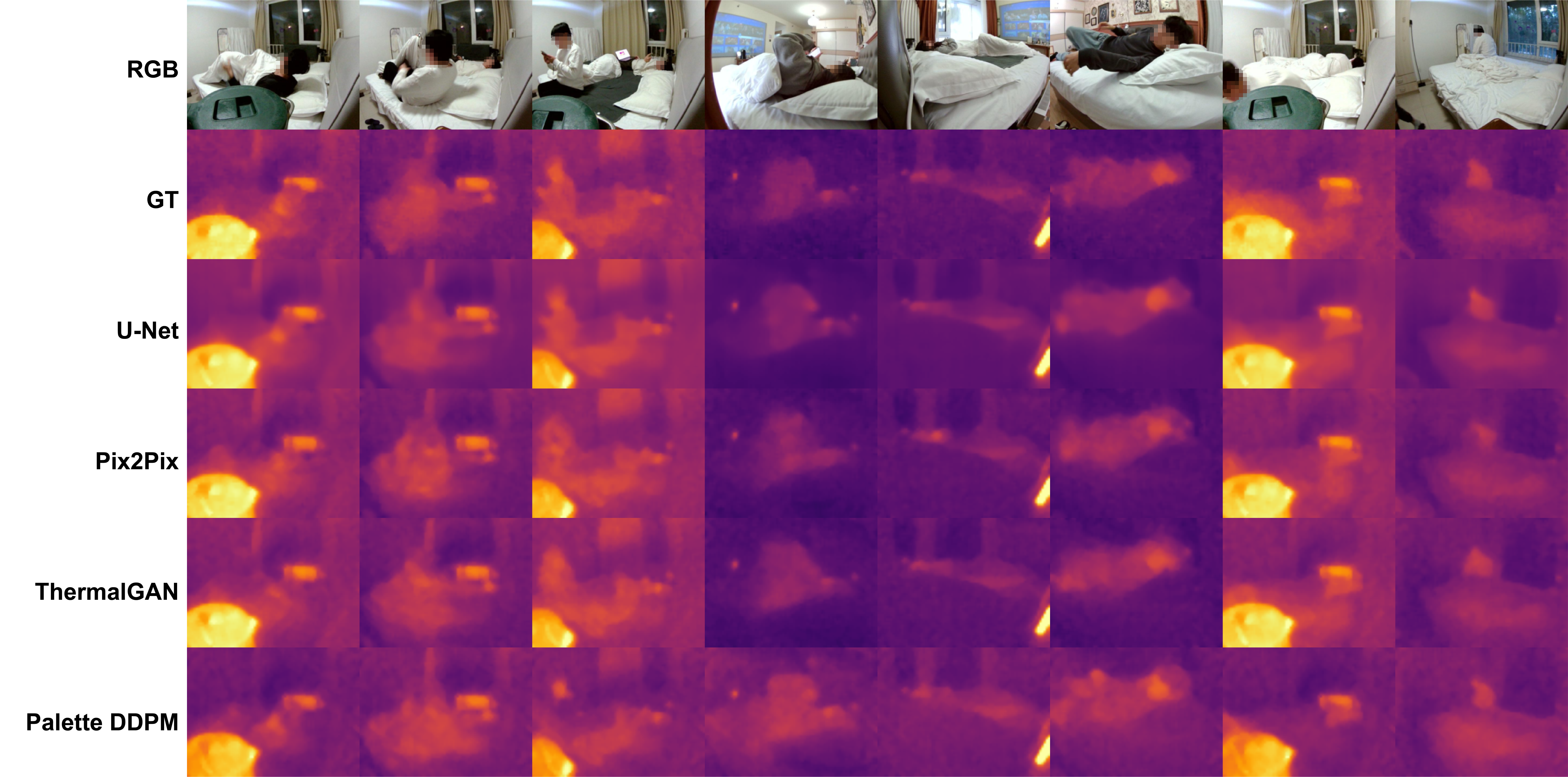}
  \vspace{-1em}
  \caption{Qualitative comparison of RGB-to-thermal translation results across different rooms.}
  \vspace{-1em}
  \label{fig:RGB2T}
\end{figure*}

Figure~\ref{fig:RGB2T} presents qualitative comparisons of RGB-to-thermal translation results across different rooms. U-Net generates thermal patterns that closely follow the spatial distribution of the ground-truth temperature fields, particularly for occupant-related high-temperature regions. Pix2Pix and ThermalGAN-style models produce visually plausible thermal representations but exhibit slight deviations in temperature intensity. Palette DDPM preserves coarse scene structures but fails to accurately recover detailed thermal distributions, which is consistent with its lower quantitative performance.

\subsubsection{Thermal-to-RGB Generation}
\label{subsec:t2rgb}

\begin{table*}[t]
\centering
\caption{Thermal-to-RGB synthesis performance. Higher PSNR, SSIM, and
MS-SSIM are better, whereas lower MAE, RMSE, $\Delta E$, LPIPS, and FID
are better.}
\vspace{-1em}
\label{tab:t2rgb}
\small
\setlength{\tabcolsep}{8pt}
\renewcommand{\arraystretch}{1.20}

\begin{tabular}{lcccccccc}
\toprule
Method &
PSNR$\uparrow$ &
SSIM$\uparrow$ &
MS-SSIM$\uparrow$ &
MAE$\downarrow$ &
RMSE$\downarrow$ &
$\Delta E\downarrow$ &
LPIPS$\downarrow$ &
FID$\downarrow$ \\
\midrule

Pix2Pix \cite{isola2017image}
& 18.590
& 0.700
& 0.785
& \textbf{0.062}
& \textbf{0.118}
& \textbf{6.700}
& 0.183
& 28.860 \\

NAFNet \cite{chu2022nafssr}
& 16.030
& 0.437
& 0.512
& 0.110
& 0.163
& 10.960
& 0.540
& 65.590\\

ControlNet-SDXL \cite{9878449}
& 10.520
& 0.398
& 0.186
& 0.237
& 0.302
& 22.830
& 0.701
& 68.790\\

ControlNet \cite{zhang2023adding}
& 10.220
& 0.333
& 0.207
& 0.252
& 0.324
& 24.220
& 0.738
& 78.280\\

BBDM \cite{li2023bbdm}
& \textbf{19.470}
& \textbf{0.797}
& \textbf{0.801}
& 0.069
& 0.133
& 7.320
& \textbf{0.161}
& \textbf{20.390}\\

\bottomrule
\end{tabular}
\end{table*}

Given a thermal image, the thermal-to-RGB synthesis task aims to generate a corresponding RGB representation while preserving the spatial configuration and occupant-related information of the input scene. We benchmark five representative image synthesis methods spanning adversarial, restoration, and diffusion-based paradigms: Pix2Pix, NAFNet, ControlNet based on Stable Diffusion~1.5, ControlNet based on SDXL, and BBDM. 
PSNR, SSIM, and MS-SSIM measure structural similarity with the paired RGB targets; MAE, RMSE, and $\Delta E$ quantify pixel and color consistency; and LPIPS and FID evaluate perceptual similarity and distribution-level realism.

In Table~\ref{tab:t2rgb}, Thermal-to-RGB synthesis exhibits different performance characteristics across pixel-level reconstruction and perceptual quality metrics. BBDM achieves the strongest overall structural and perceptual performance, obtaining the highest PSNR (19.470), SSIM (0.797), and MS-SSIM (0.801), together with the lowest LPIPS (0.161) and FID (20.390).
Pix2Pix achieves the best pixel-level and color consistency with the paired RGB targets. It obtains the lowest MAE (0.062), RMSE (0.118), and $\Delta E$ (6.700) among all evaluated methods, demonstrating stronger local correspondence between the synthesized and ground-truth RGB images. Compared with BBDM, Pix2Pix provides more accurate pixel reconstruction but achieves slightly lower structural and perceptual similarity, suggesting a trade-off between pixel fidelity and perceptual quality. This T2RGB generation cannot identify personal information.



\section{Real-World Deployment and Operation}
\label{sec:deployment}

\begin{figure*}[tp]
  \centering
  \includegraphics[width=0.7\textwidth]{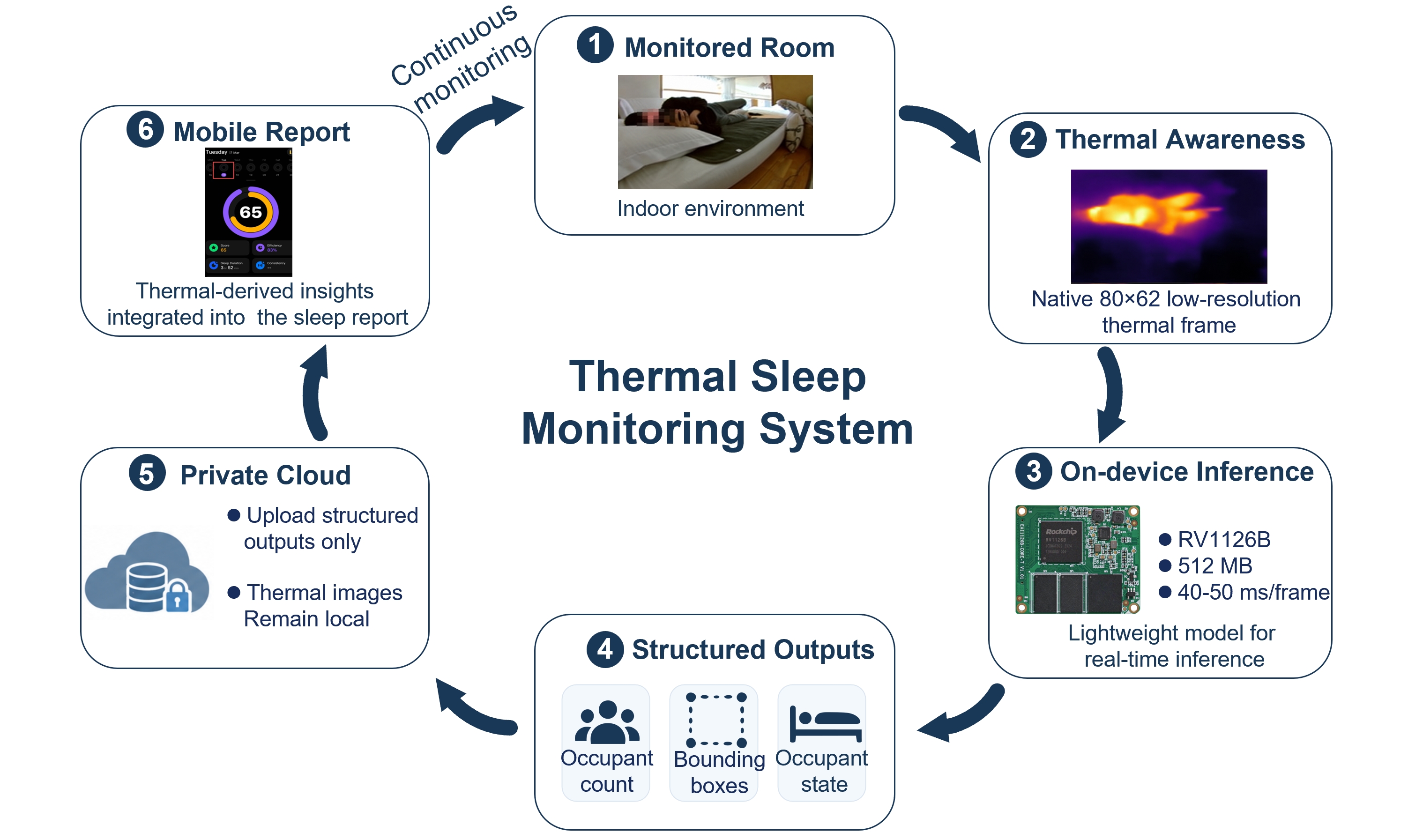}
  \vspace{-1em}
  \caption{Overview of the real-world thermal sleep monitoring system, including thermal awareness, on-device inference, structured output transmission, cloud-side integration, and mobile sleep reporting.}
  \vspace{-1em}
  \label{fig:real}
\end{figure*}

We deployed the low-resolution thermal awareness module in a commercial sleep-monitoring platform in Fig.~\ref{fig:real}. The module processes native $80\times62$ thermal frames and outputs occupant count, occupant bounding boxes, and occupancy states. These outputs are used by downstream functions such as occupancy estimation, occupant-state tracking, and occupied time (occupancy detection). We characterize the deployed system from three perspectives: perception performance, on-device inference, and privacy preservation.

First, the deployment relies on the fact that low-resolution thermal observations retain sufficient information for reliable occupancy perception. This is supported by the benchmark results presented in the previous section, where the thermal modality achieves stable performance for occupancy detection, state recognition, and occupancy estimation. Thermal awareness also remains effective under very low illumination or complete darkness without relying on visible imaging. We use it for continuous sleep monitoring in bedrooms.

Second, all model inference is performed locally on the device. We deployed a lightweight YOLO series detection model on an RV1126B edge processor with 512~MB memory using the RKNN framework with FP16 inference. After model and deployment optimization, the system achieves an inference latency of approximately 40--50~ms per thermal frame. When all on-device functions are enabled, the overall device power consumption is approximately 2.8- 3.5~W. This configuration enables continuous real-time perception without relying on remote servers.

Third, the deployment architecture is designed to limit privacy exposure at both the sensing and data transmission stages. Compared with RGB images, low-resolution thermal frames contain substantially fewer fine-grained visual details that could reveal personal appearance. In addition, raw thermal frames are processed locally and remain on the device. Only structured inference outputs, including occupant count, bounding boxes, and occupant states, are transmitted via Wi-Fi to a private cloud backend for downstream system processing. This design avoids continuous transmission of raw thermal data while retaining the information required for monitoring and reporting.

To evaluate the robustness of the deployed thermal perception module under real-world residential conditions, we assessed its performance through extended field validation and large-scale commercial operation.

\textbf{Field validation.} 
The system underwent extended field validation in 100 residential bedrooms across 100 households in Beijing from December 2025 to September 2026. We recruited 100 volunteers who were concerned about their sleep quality but were unwilling to use contact-based sleep monitoring devices, such as smartwatches or rings. Each participant deployed the proposed system in their own bedroom for the field validation. This stage evaluated system operation under uncontrolled residential conditions and identified deployment-specific error sources that were not captured during offline evaluation. Overall, satisfactory system operation was reported in 96 of the 100 households (96\%). An interesting failure mode emerged during winter: in 4 households (4\%), vertically installed radiators produced thermal patterns resembling an off-bed occupant, resulting in false detections. This failure mode had not been observed during controlled evaluation. The system underwent approximately one month of additional refinement to improve robustness against such environmental thermal interference. These field trials therefore enabled deployment-specific failure modes to be identified and addressed before broader deployment.

\textbf{Commercial operation.}
The system entered commercial operation in June 2026. Users simply place the product beside the bed, similar to a bedside lamp, without the need to wear or install contact-based devices. To protect user privacy, the product is equipped with a low-resolution thermal camera without an RGB camera. During the first three months, 1,000 devices were delivered to users across multiple countries and regions, including China, the United States, Singapore, Germany, France, the United Kingdom, Malaysia, and Israel. User feedback was collected for all 1,000 deployed devices. Thermal perception-related issues were reported in 30 cases, whereas no thermal-related issues were reported in the remaining 970 cases.

 Analysis of the 30 thermal-related reports identified three dominant error sources. Improper device placement accounted for 15 cases (50.0\%), primarily by altering the effective sensing geometry. Solar heating accounted for 10 cases (33.3\%), particularly in the afternoon, when heated objects or residual thermal patterns could trigger false detections. The remaining 5 cases (16.7\%) were associated with multi-occupant occlusion, where closely positioned occupants could be merged into a single detection.




The real-world evaluation revealed deployment-specific challenges that were not fully captured by offline benchmark performance. Taken together, field validation and large-scale commercial operation provide complementary evidence for the practical feasibility, operational robustness, and scalability of native low-resolution thermal perception in privacy-sensitive indoor environments. 













\section{Limitations}
\label{sec:Limitations}

Several limitations remain for future research. First, although the dataset covers diverse indoor rooms and occupant conditions, extending data collection to additional environments, such as healthcare facilities and assisted-living scenarios, would further evaluate generalization. Second, the current benchmark mainly focuses on frame-level perception, while temporal modeling of occupant behaviors may improve robustness under ambiguous thermal observations. 

\section{Conclusion}
\label{sec:conclusion}

This work presents an RGB--thermal benchmark for privacy-preserving indoor human perception using low-resolution indoor thermal images. We introduce a large RGB--thermal dataset for indoor human perception, covering 20 rooms and containing 87,016 temporally matched RGB--thermal pairs and 115,557 annotated occupant bounding boxes. We evaluate indoor human state detection and recognition, occupancy estimation, RGB--thermal translation, and new thermal-RGB generation tasks in unseen indoor environments. By validating practical feasibility through long-term field testing and commercial operation with high sales volume and an low reported failure rate, our deployed sleep monitoring system demonstrates high accuracy and robustness, as well as practical application value. Our practical deployment reveals key challenges and thus highlights future research directions, including environmental thermal interference, limited sensing range, and multi-person occlusion.






\section*{Acknowledgment}
This work was supported by the National Natural Science Foundation of China under Grant 62192751, partly by the 111 International Collaboration Program of China under Grant BP2018006, and partly by the BNRist Program under Grant No. BNR2019TD01009, and in part by China National Innovation Center of High-Speed Train Project Under Grant CX/KJ-2020-0006.


\bibliographystyle{IEEEtran}

\bibliography{cas-refs}




\end{document}